\documentclass{article}
\usepackage{spconf,amsmath,graphicx}
\usepackage[usenames,dvipsnames]{color} 
\usepackage{algpseudocode}
\usepackage{algorithm}

\title{Visible and infrared self-supervised fusion trained on a single example}
\name{Nati Ofir and Jean-Christophe Nebel}

\address{Kingston University London}

\begin{document}
	%
	\maketitle

	\begin{abstract}
        Multispectral imaging is an important task of image processing and computer vision, which is especially relevant to applications such as dehazing or object detection. With the development of the RGBT (RGB \& Thermal) sensor, the problem of visible (RGB) to Near Infrared (NIR) image fusion has become particularly timely. Indeed, while visible images see color, but suffer from noise, haze, and clouds, the NIR channel captures a clearer picture. The proposed approach fuses these two channels by training a Convolutional Neural Network by Self Supervised Learning (SSL) on a single example. For each such pair, RGB and NIR, the network is trained for seconds to deduce the final fusion. The SSL is based on the comparison of the Structure of Similarity and Edge-Preservation losses, where the labels for the SSL are the input channels themselves. This fusion preserves the relevant detail of each spectral channel without relying on a heavy training process. Experiments demonstrate that the proposed approach achieves similar or better qualitative and quantitative multispectral fusion results than other state-of-the-art methods that do not rely on heavy training and/or large datasets.

        \end{abstract}
	
\section{Introduction} \label{sec:intro}

Multispectral image fusion is a well-studied area with many applications. For example, while the visible RGB channel $(0.4-0.7\mu m)$ perceives color information, the Near-Infrared (NIR) one $(0.8-2.5\mu m)$ sees beyond haze and fog and suffers less from noise in low-light conditions. Since each spectral channel captures different information about a scene, their fusion is particularly informative providing additional details.

On one hand, deep learning-based fusion approaches, such as attention-based \cite{li2020nestfuse}, deliver high-quality results, but require a lengthy training phase which prevents real-time learning. On the other hand, classic image fusion techniques do not rely on any training and can produce results in an instant \cite{ofir2023multispectral}. However, their drawback is in their lower ability to preserve input details, as shown in the experiments section of this paper. Ideally, one would like an approach that can produce high-quality multispectral fusion in real-time without requiring either heavy pre-training or a large training dataset. Given such a solution, a single image fusion could be trained per request by image editing software. Thus, any multispectral camera could feature a fusion channel allowing users to see the combined details captured by each spectrum without having to flicker between the images captured by each channel. 

Herein, exploiting Deep Learning (DL) technology, a novel method is proposed that delivers both state-of-the-art quality and speed. This is achieved by designing an architecture based on a Convolutional Neural Network (CNN), in which weights are learned for each fusion pair using a single example. Eventually, high-quality fusion is produced in a self-supervised learning (SSL) manner such that no manual labeling is required. As very few training data are used, learnable parameters can be calculated in only a few seconds using a recent standard Graphic Processing Unit (GPU). 

\begin{figure}
	\centering
	\includegraphics[width=75px]{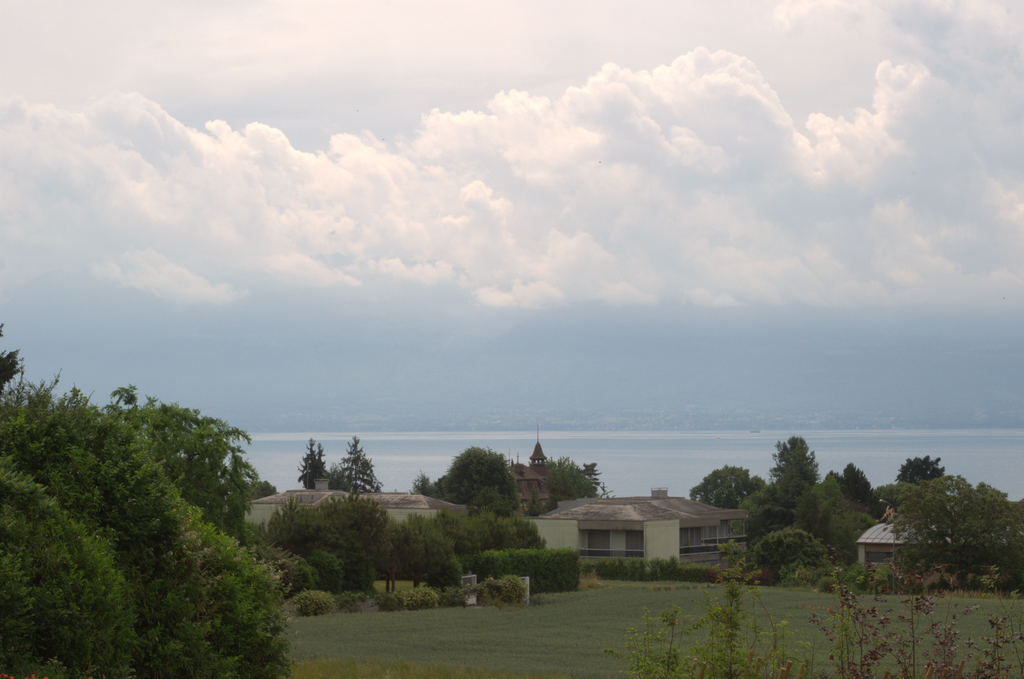}~
	\includegraphics[width=75px]{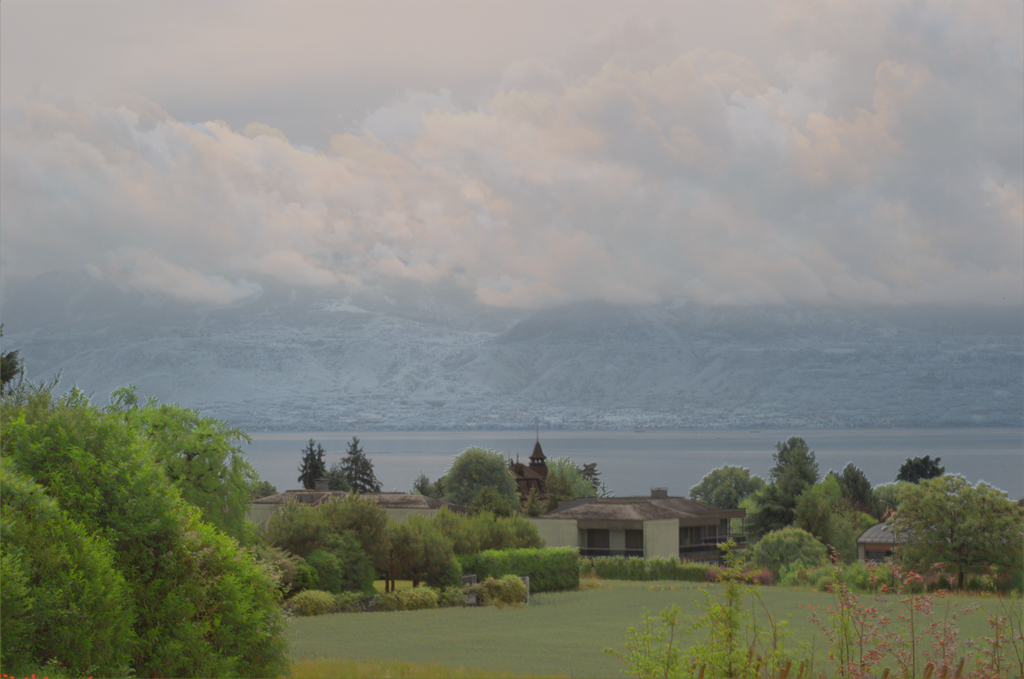}~
	\includegraphics[width=75px]{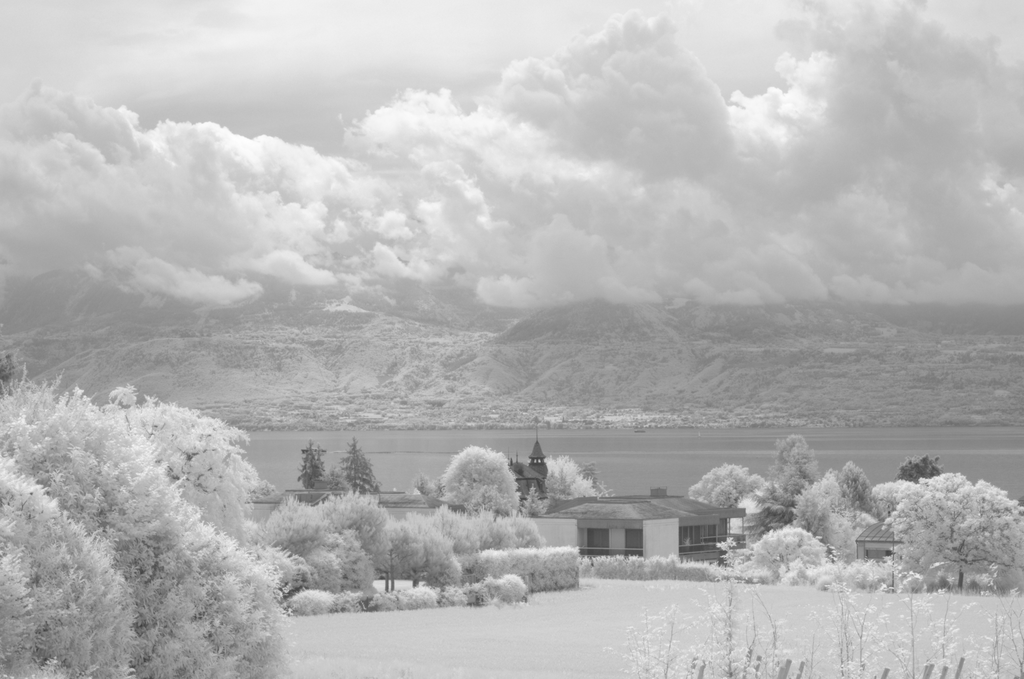}
	\caption{Multispectral image fusion using RGB and NIR channels from the 'country' category from the VIS-NIR dataset \cite{BS11}. Left: Input RGB channel. Middle: Fusion outcome. Right: Input NIR channel.}
	\label{fig:fusion}
\end{figure}

Figure \ref{fig:fusion} shows an example of the outcome of the proposed method when fusing RGB and NIR images. This result combines while maintaining, the information from both inputs. Indeed, not only the color information from the RGB sensor is preserved, but also the far mountains that could only be seen in NIR are now visible. Even though this method relies on learned artificial intelligence, the outcome is natural looking without unrealistic artifacts.

Despite the development of RGBT (RGB \& Thermal) cameras, there are many instances when the input channels are not aligned with each other, for example, small misalignments are found in popular datasets \cite{BS11}. Thus, multispectral image registration may be required as a preprocessing step. Although images could be aligned in advance by methods tailored for multispectral imaging either DL-based \cite{ofir2018deep} or traditional computer vision-based \cite{ofir2018registration}, an elegant solution is suggested as it can be integrated into the proposed CNN architecture: a Spatial-Transformation-Network (STN) \cite{jaderberg2015spatial} is learned in a holistically end-to-end way to compute the final aligned fusion results. As shown in Figure \ref{fig:fusion}, where the two channels were slightly misaligned, the CNN output does not suffer from channel misregistration.

This manuscript is organized as follows. Section \ref{sec:previous} covers the previous methods for image fusion and registration. Next, in Section \ref{sec:fusion} the proposed approach is detailed including the CNN-based architecture, the training algorithm, and its loss functions. Then, Section \ref{sec:results} compares fusion performance with other methods that do not rely on a time-consuming training phase. Finally, this paper is concluded in Section \ref{sec:conclusions}.

\section{Previous work} \label{sec:previous}

Image fusion is a standard problem of image processing and computer vision that was initially addressed using classic approaches. Early methods utilized signal characteristics such as Wavelets-based method \cite{chipman1995wavelets}. Later, to overcome multi-focus image capturing, laplacian pyramid blending was proposed  \cite{wang2011multi}. Principal Component Analysis (PCA) was also considered to extract statistical features of the input images \cite{patil2011image}. More recently, spectral analysis of these images was introduced to enhance their fusion \cite{ofir2018registration}. Finally, superpixels \cite{achanta2012slic} segmentation was exploited for content-based multispectral fusion \cite{ofir2023multispectral}. However, with the DL revolution, many novel solutions have been produced delivering state-of-the-art blending performances \cite{li2018infrared}. Consequently, a fusion of RGB and NIR images has become a powerful and practical solution to enhance DL-based object detection \cite{liu2022target}. 

The proposed method utilizes DL techniques and architecture, however, does not depend on heavy training processes and large datasets contrary to the most recent approaches. Indeed, the creation of a customized model on the fly can be extremely valuable for dedicated image editing software and applications. Inspired by performance of CNNs trained on a single example for super-resolution \cite{shocher2018zero} and image-generation by Generative-Adverserial-Network (GAN)\cite{shaham2019singan}, this work is, to the best of our knowledge, the first to utilize single-image training for multispectral image fusion.

If the input spectral channels are not geometrically aligned, an initial step of multispectral registration is required. Although single-channel registration can be carried out by engineered feature descriptors like Scale-Invariant-Feature-Transform (SIFT) \cite{lowe2004distinctive}, such alignment methods usually fail with multispectral channels. Therefore, an approach tailored to such a scenario is needed. Whereas a descriptor that is invariant to different spectra could be based on edge detection \cite{ofir2018registration}, like Canny \cite{canny1986computational}, it would underperform in the presence of geometric transformations. To address this, usage of mutual information-based registration has proved beneficial as it can usually handle translations and small optical flow fields \cite{maes1997multimodality}. Eventually, the most promising methods rely on DL. They include computing spectra-invariant descriptors, even if they have geometrical constraints \cite{ofir2018deep}. Learning a hybrid network for multispectral key point matching has proved to deliver much better accuracy \cite{baruch2021joint}. However, the main drawback is that it relies on a training dataset that must be manually labeled, which is often not practical. Thus, multispectral image alignment remains a challenging problem that has not been solved yet.

Although the development of RGBT cameras \cite{li2020challenge} should make the need for registration obsolete, they are not yet widespread and many benchmarking data sets, such as the VIS-NIR dataset \cite{BS11}, contain small misalignments between images. Thus, in practice, the multispectral registration step remains needed. As regular registration methods fail for RGB-IR alignment and there is a lack of suitable training datasets, it is proposed to address multispectral misalignments by solving them holistically using the learned CNN. Indeed, the geometric correction can be trained using a spatial transformation network \cite{jaderberg2015spatial}, which computes geometric transformations by end-to-end learning. 

Many DL methods, such as contrastive learning \cite{jaiswal2020survey}, are self-supervised, allowing CNN training without the existence of labels. As being independent of human labeling is a requirement that is mandatory for practical multispectral image fusion, SSL is the approach of choice. In this paper, the proposed method uses the input spectral channels as a label for their fusion by taking advantage of Structure of Similarity Measure (SSIM) \cite{lu2019level} and Edge Preservation (EP) losses \cite{ofir2021multi}. As a whole, this study introduces a holistic solution for visible-to-infrared fusion and registration based on SSL.

\section{The proposed multispectral fusion} \label{sec:fusion}
This section introduces a novel method to fuse visible and infrared multispectral images by training a fusion CNN on a single example using self-supervised loss functions.

\subsection{Network architecture}
\begin{figure}
	\centering
	\includegraphics[width=200px]{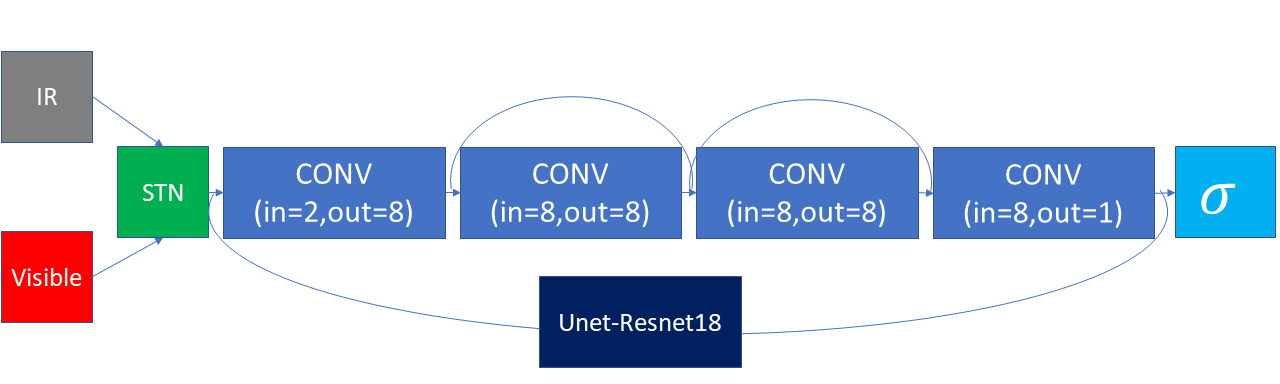}
	\caption{CNN architecture of the proposed method. The network inputs two channels and outputs a single channel of image fusion. Following image alignment using an STN, four convolutions with two skip connections are applied. In addition, a UNet-Resnet18 is trained in parallel to compute an accurate fusion map to enhance quality.}
	\label{fig:architecture}
\end{figure}

The proposed CNN architecture for image fusion processes two images of any dimension from different channels to output a single combined image of the same height and width as the input, see Figure \ref{fig:architecture}. The overall network is holistically designed such that an image fusion can be computed very quickly by self-supervised training on a GPU. 

The compact fusion network contains four convolutions of kernel 3x3, the first three are followed by $ReLU(x) = max(x,0)$ activation and the final output-convolution is followed by $Sigmod(x) = \frac{e^x}{1+e^x}$. The architecture contains two skip connections that are based on numeric addition. Before the feed-forward CNN, an off-the-shelf STN is applied to align the spectral channel. In addition, a UNet \cite{siddique2021u} with Resnet18 backbone \cite{tai2017image} is applied in parallel to the feed-forward CNN, to get a smooth fusion with semantic information.
In order to keep the CNN versatile and allow fast training, the designed architecture relies on only $\approx 14M$ learnable parameters. Details about the whole CNN parameters are shown in Table \ref{table:architecture}. 

\begin{figure}
	\centering
	\includegraphics[width=200px]{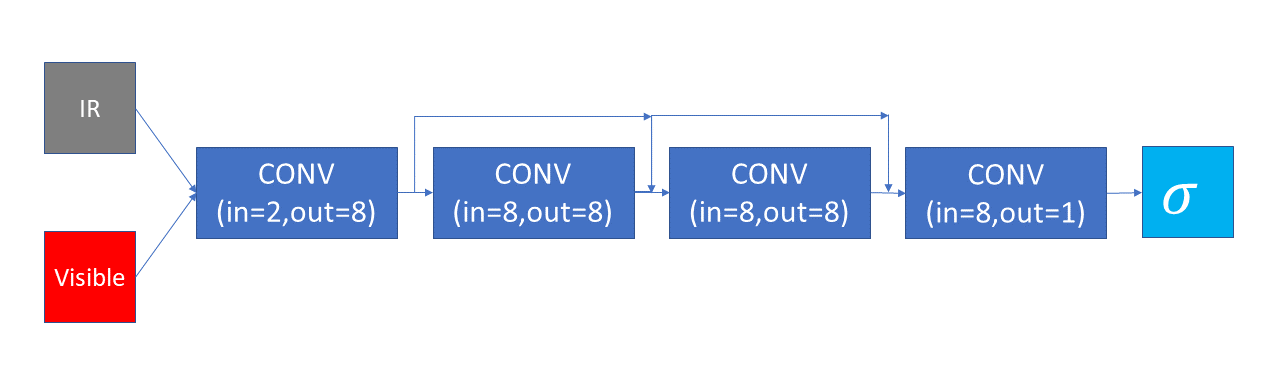}
	\caption{Compact CNN architecture used in the ablation study reported in Section \ref{sec:results}.}
	\label{fig:compact}
\end{figure}

Although a compact fusion could be performed by using only the convolution layers, see Figure \ref{fig:compact}, an ablation study is conducted in Section \ref{sec:results} to evaluate the contribution of each component of the architecture to the final fusion results.

\begin{table}[htb]
	\centering
	\begin{tabular}{| l  | c | c | c |}
        \hline
        \hline
		Layer  & In Ch. & Out Ch. & Operation \\
        \hline
        STN  & 2 & 2 & Affine \\ 
        UNet-Resnet18 & 2 & 1 & Unet(STN) \\ 
    	conv1  & 2 & 8 & ReLU(STN)  \\
        conv2  & 8 & 8 & ReLU(conv1) \\
        add1  & - & - & conv1+conv2 \\
        conv3  & 8 & 8 & ReLU(add1) \\
        add2 & - & - & add1+conv3 \\
		conv4  & 8 & 1 & Sigmoid(add2+Unet) \\
		\hline
		\hline
	\end{tabular}
	\caption{Layers of the fusion CNN architecture trained in parallels to the UNet-Resnet18 fusion.}
	\label{table:architecture}
\end{table}

\subsection{Training algorithm}

To train the CNN-based method, a training loop algorithm is introduced, see Algorithm \ref{alg:train}. First, the IR and RGB input images are loaded and the RGB image is converted into a gray image 'GRAY'. Second, the CNN is initialized with random weights. Then, iteratively the training loop computes the CNN weights to fuse the specific pair of NIR and GRAY images. During this process the network's weights are updated taking advantage of a combination of SSIM \cite{ssim} and Edge Preservation \cite{ofir2021multi} losses. Finally, after completion of the training loop, the fused gray image is computed and used to adjust the RGB channel image to generate the RGB fusion image. Note that for simplicity the fusion is performed on the GRAY channel and the color is added in a post-processing step.

\begin{algorithm}
\caption{Visible and infrared self-supervised fusion}\label{alg}
\begin{algorithmic}
\State $Fusion(IR, RGB):$
\Require $IR \gets InfraredImage$
\Require $RGB \gets VisibleImage$
\State $GRAY \gets rgb2gray(RGB)$
\State $CNN \gets InitNetwork()$
\State $epoch \gets 0$
\While{$epoch \leq MaxEpoch$}
\State $Fusion \gets CNN(IR, RGB)$
    \State $SsimLoss \gets ssim(Fusion,IR)+edge(Fusion,RGB)$
    \State $EdgeLoss \gets edge(Fusion,IR)+edge(Fusion,RGB)$
    \State $loss \gets SsimLoss+EdgeLoss$
    \State $CNN \gets BackPropogation(CNN, loss)$
\EndWhile
\State $GrayFusion = CNN(IR, RGB)$
\State $RgbFusion = \frac{GrayFusion}{GRAY} \cdot RGB$
\end{algorithmic}
\label{alg:train}
\end{algorithm}

\subsection{Loss functions}

The loss functions used to train the CNN are SSIM and EP, where each of them is self-labeled using the input images.

As the similarity function, SSIM is correlated to the human visual system and has a differentiable loss definition \cite{huang2020deep}, it is widely used for understanding the perception of similar images. Given two input images $I_1$ and $I_2$, the SSIM is defined by:
$\frac{(2\mu_1\mu_2+c_1)(2\sigma_{12}+c_2)}{(\mu_1^2+\mu_2^2+c1)(\sigma_1^2+\sigma_2^2+c_2)}$,
where $\mu$ is the mean of each image, $\sigma$ is the standard deviation and $\sigma_{12}$ is the joint covariance. 
Regarding the Edge Preservation loss, it is a standard reconstruction loss, applied after image gradient detection.
$EP(I_1,I_2) = ||\nabla I_1(x)-\nabla I_2(x)||_2^2$.

Note that it is shown in the experiment Section \ref{sec:results} that using the EP loss in addition to the SSIM loss improves the quantitative fusion results of the method.

\subsection{Multispectral registration}

When there are small misalignments between the spectral channels, such as in the VIS-NIR dataset \cite{BS11}, they can be aligned holistically by the various convolutions of the proposed CNN architecture. However, when the miss-registration is significant, it is proposed to add a block to the proposed self-supervised approach. The idea is to apply a spatial transformation network \cite{jaderberg2015spatial} to the NIR channel image at the start of the architecture and to train the whole network as previously described. Note that in more extreme cases of miss-registration, a more sophisticated matching solution is required, e.g., the algorithm presented in \cite{baruch2021joint}.

\section{Results} \label{sec:results}

\begin{figure}[tbh]
	\centering
	\includegraphics[width=70px]{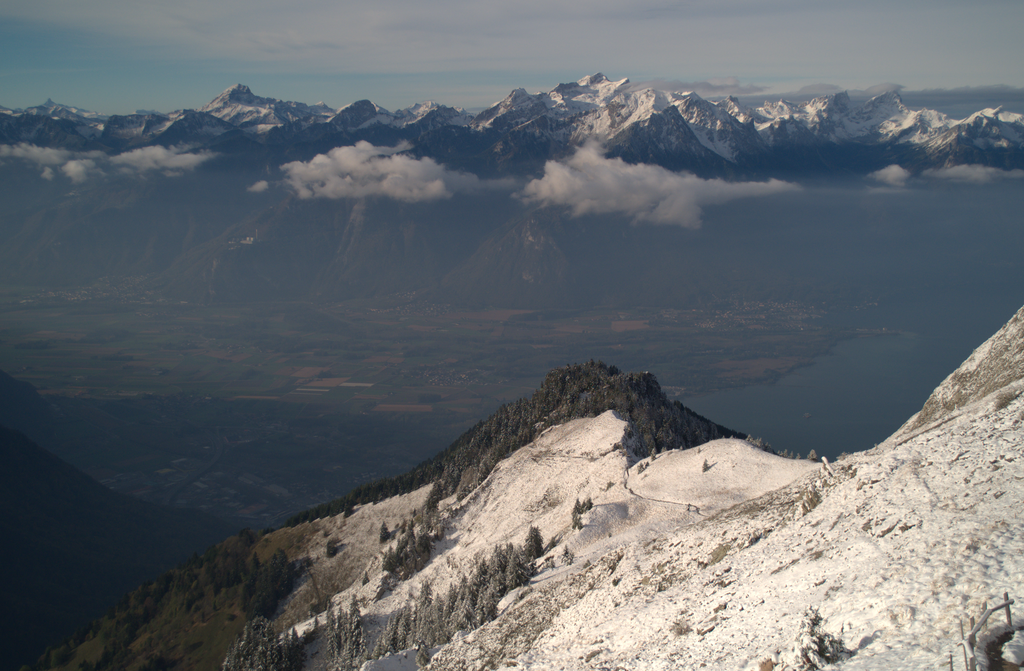}
	\includegraphics[width=70px]{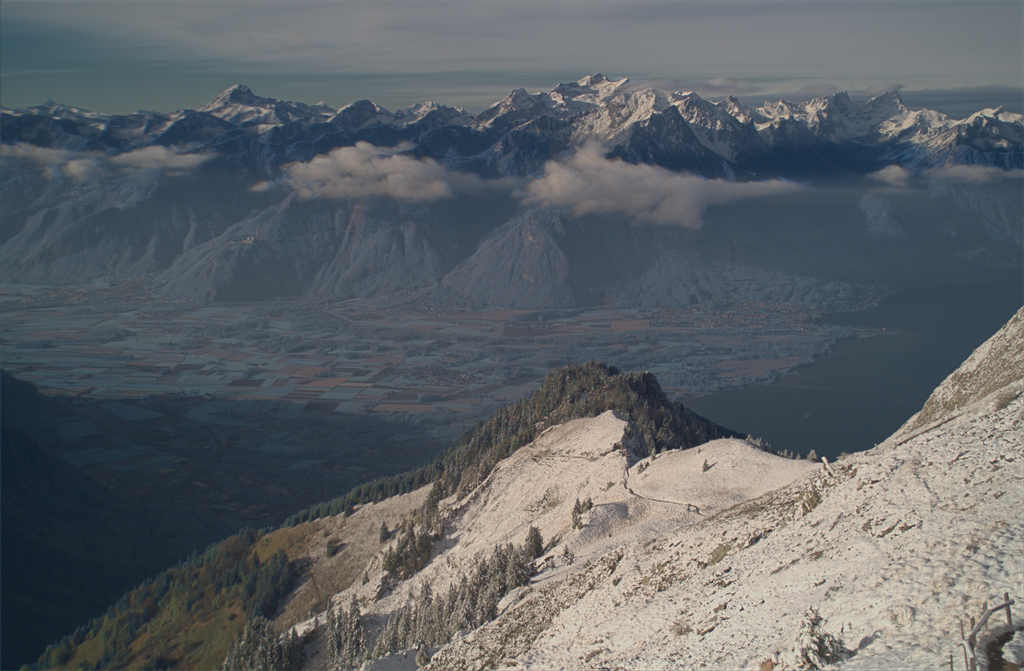}
	\includegraphics[width=70px]{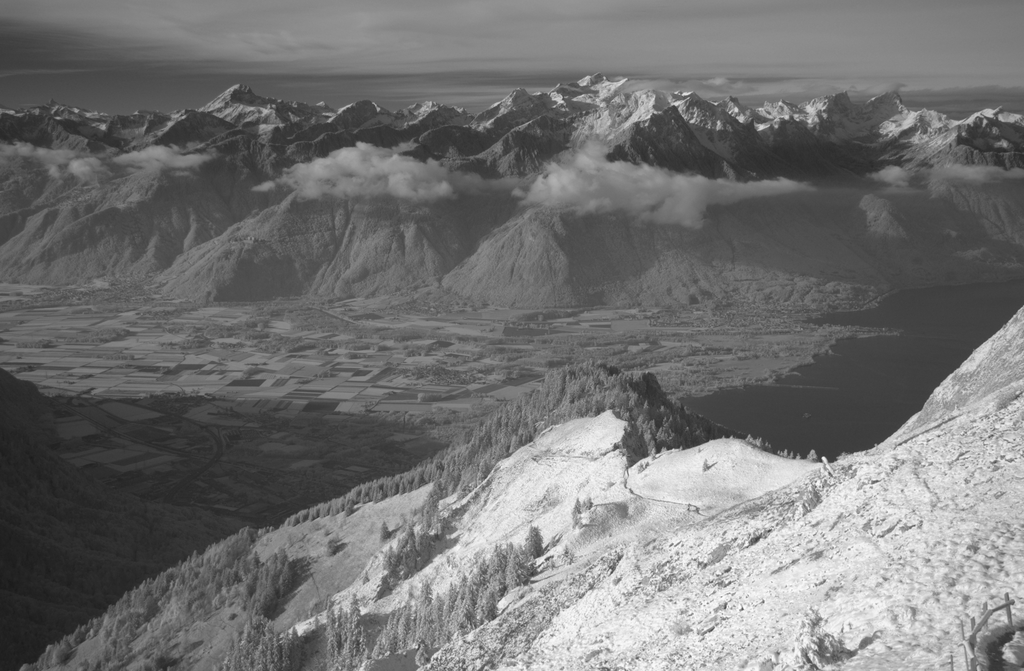}\\ [0.1cm]
	\includegraphics[width=70px]{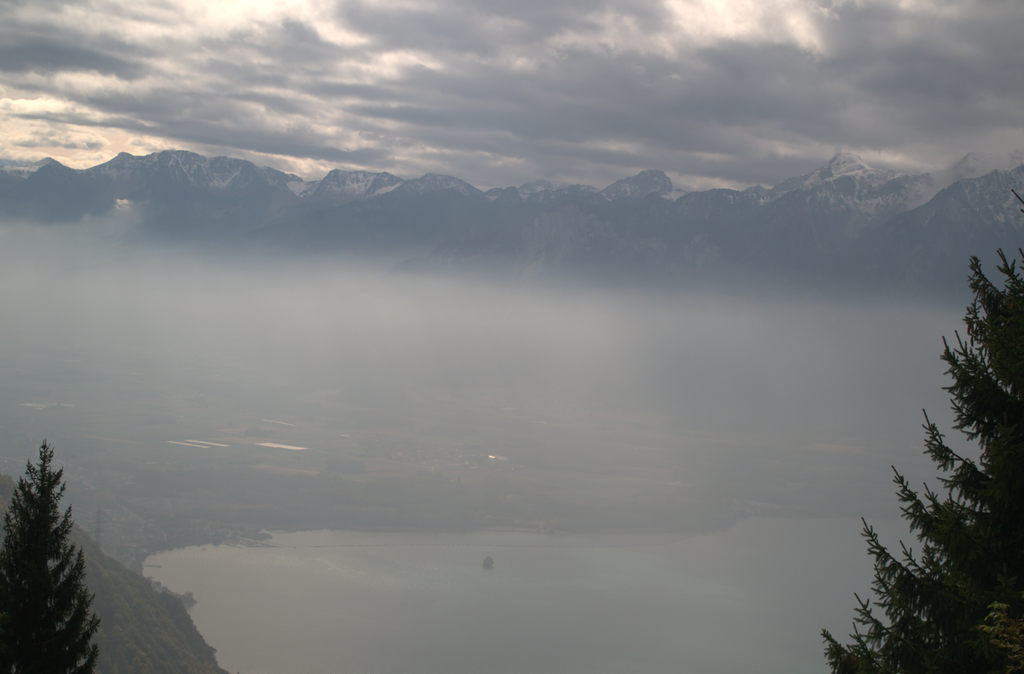}
	\includegraphics[width=70px]{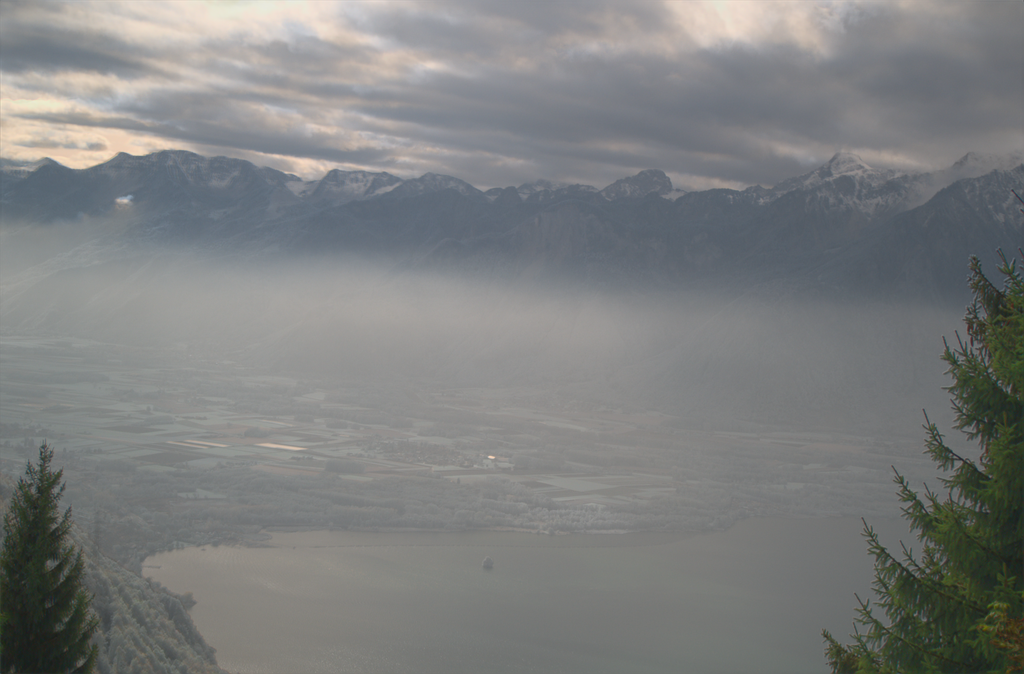}
	\includegraphics[width=70px]{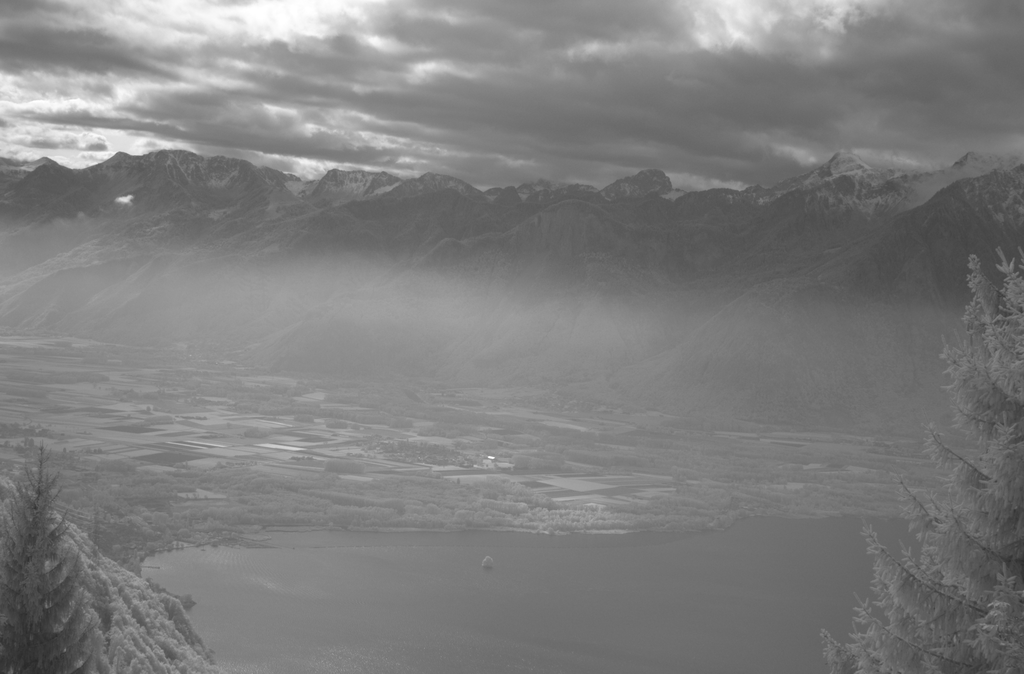}\\ [0.1cm]
	\includegraphics[width=70px]{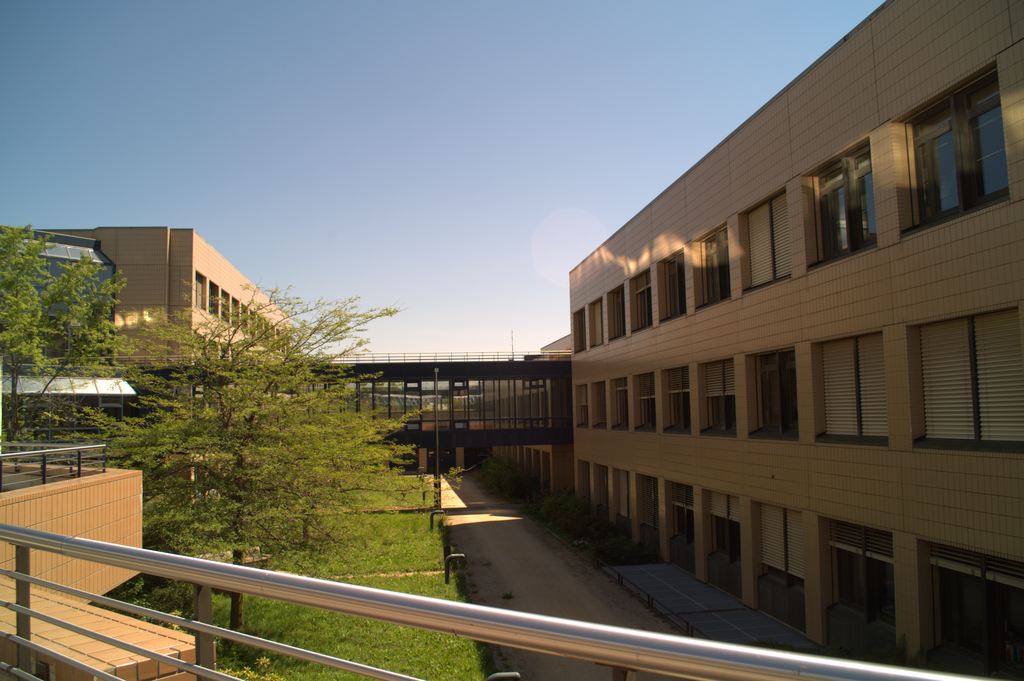}
	\includegraphics[width=70px]{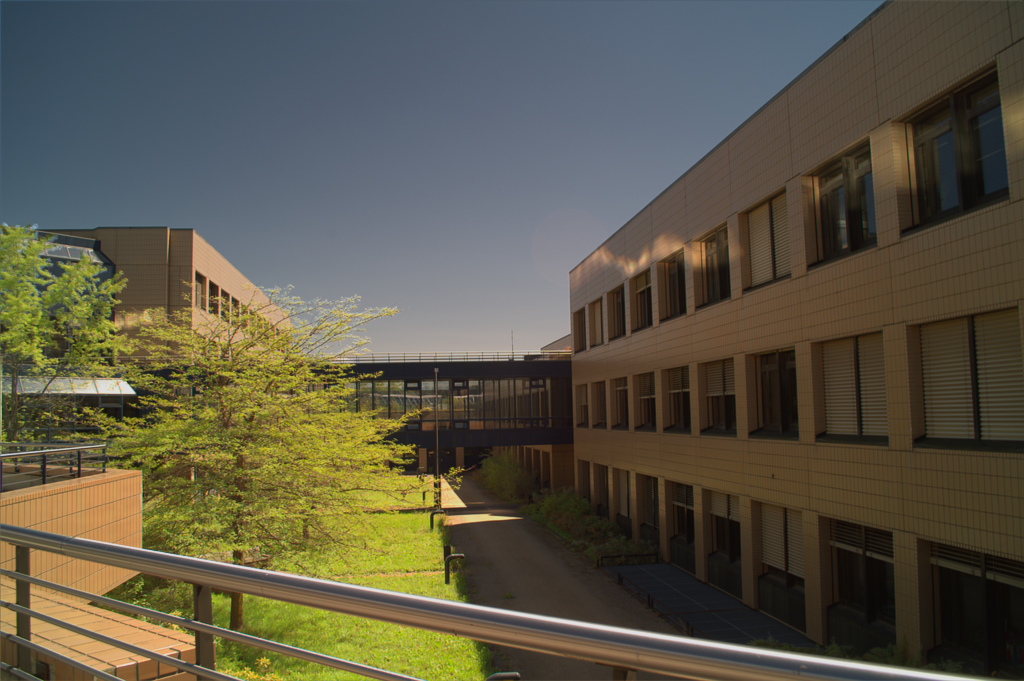}
	\includegraphics[width=70px]{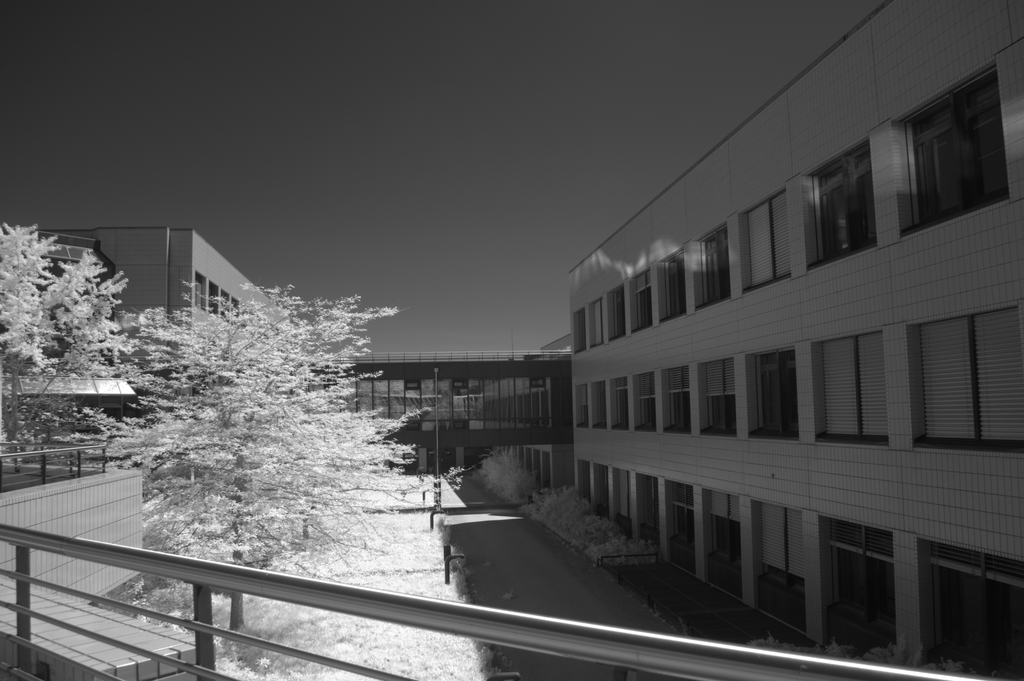}\\ [0.1cm]
	\includegraphics[width=70px]{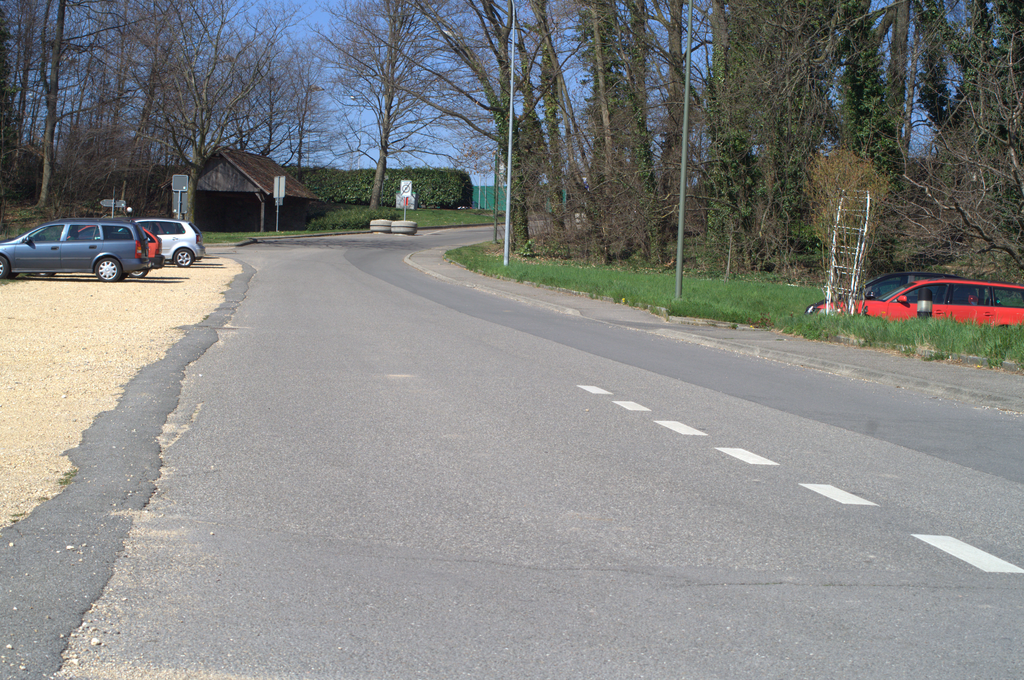}
	\includegraphics[width=70px]{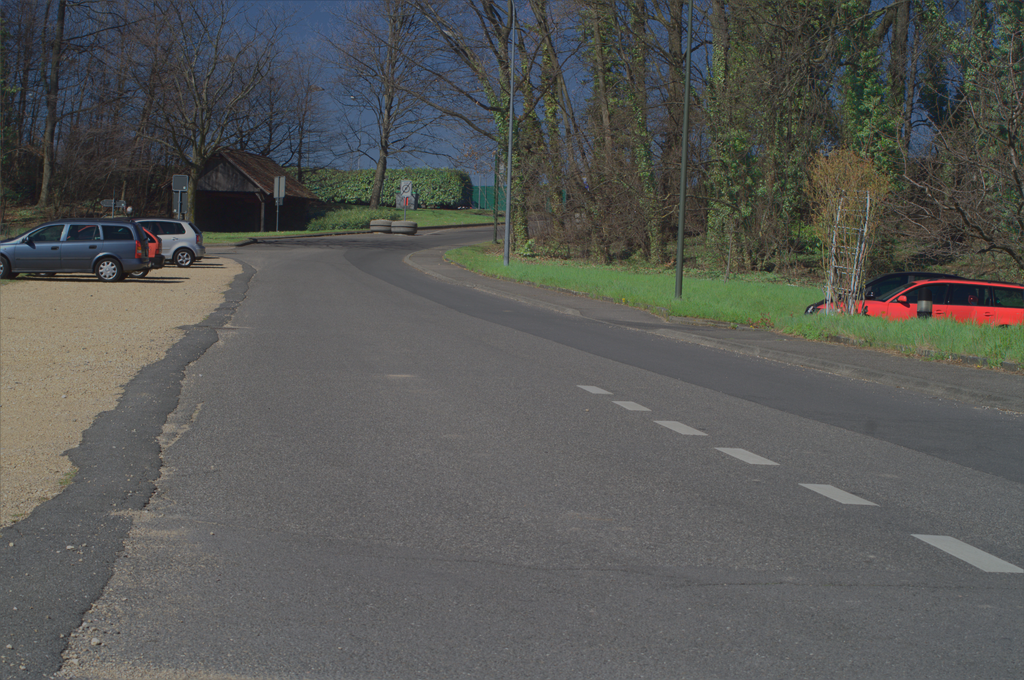}
	\includegraphics[width=70px]{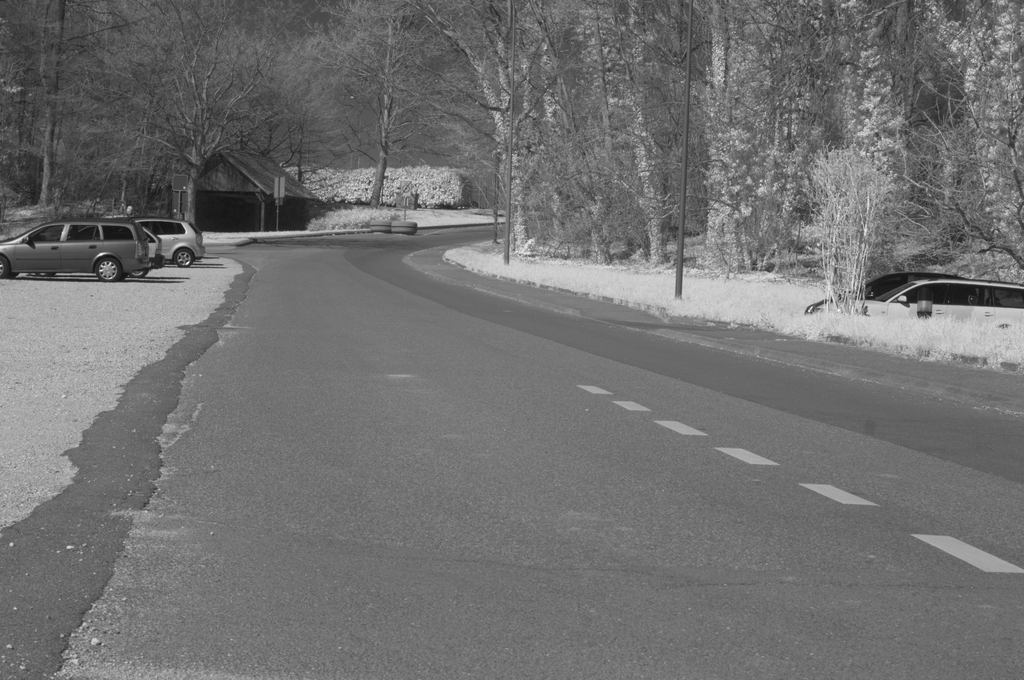}  
	\caption{Outcomes of the proposed multispectral image fusion. From left to right: input RGB,  fused, and input NIR images. From top to bottom: images from the 'Mountain', 'Country', 'Urban', and 'Street' categories of the VIS-NIR dataset.}
	\label{fig:fusion_results}
\end{figure}

\begin{figure}[tbh]
	\centering
	\includegraphics[width=80px]{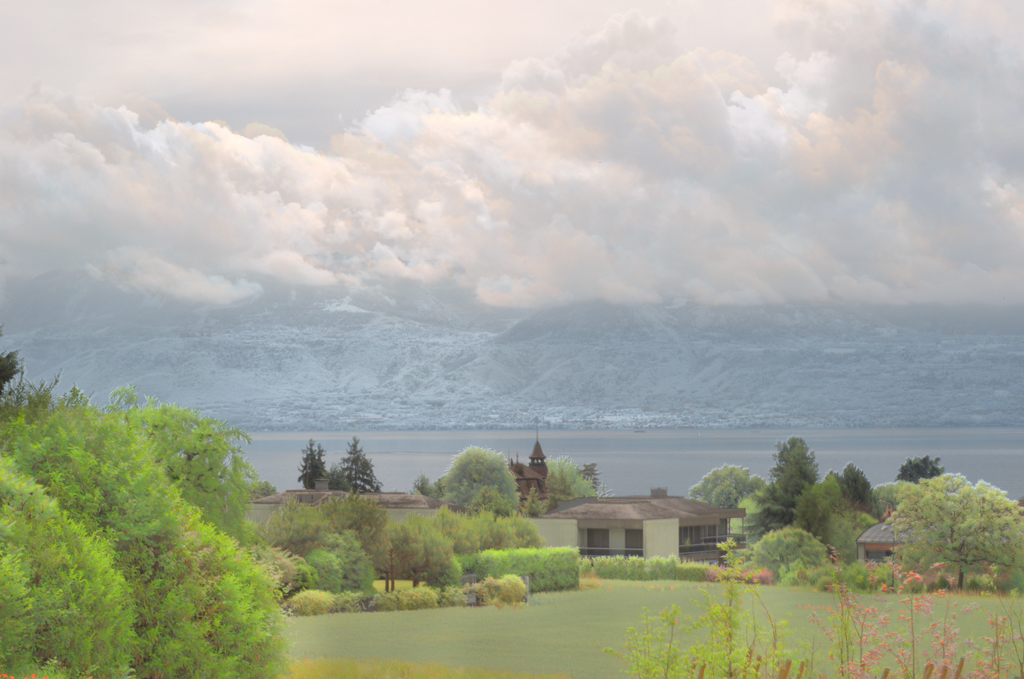}
	\includegraphics[width=80px]{deep_fusionRGB1.png}\\ [0.1cm]
	\includegraphics[width=80px]{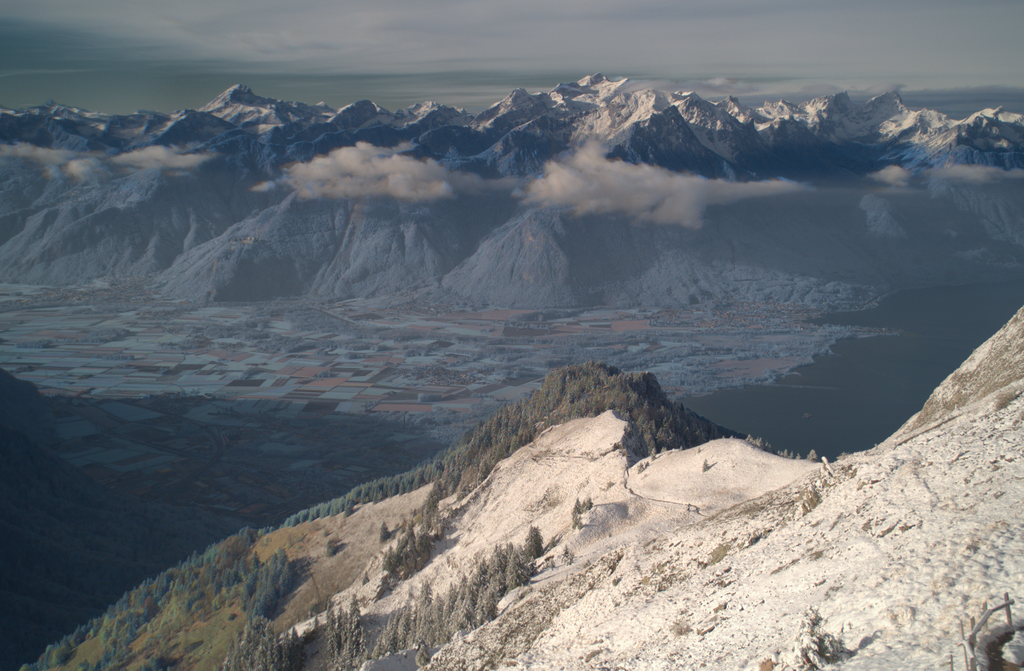}
	\includegraphics[width=80px]{deep_fusionRGB43.png}\\ [0.1cm]
	\includegraphics[width=80px]{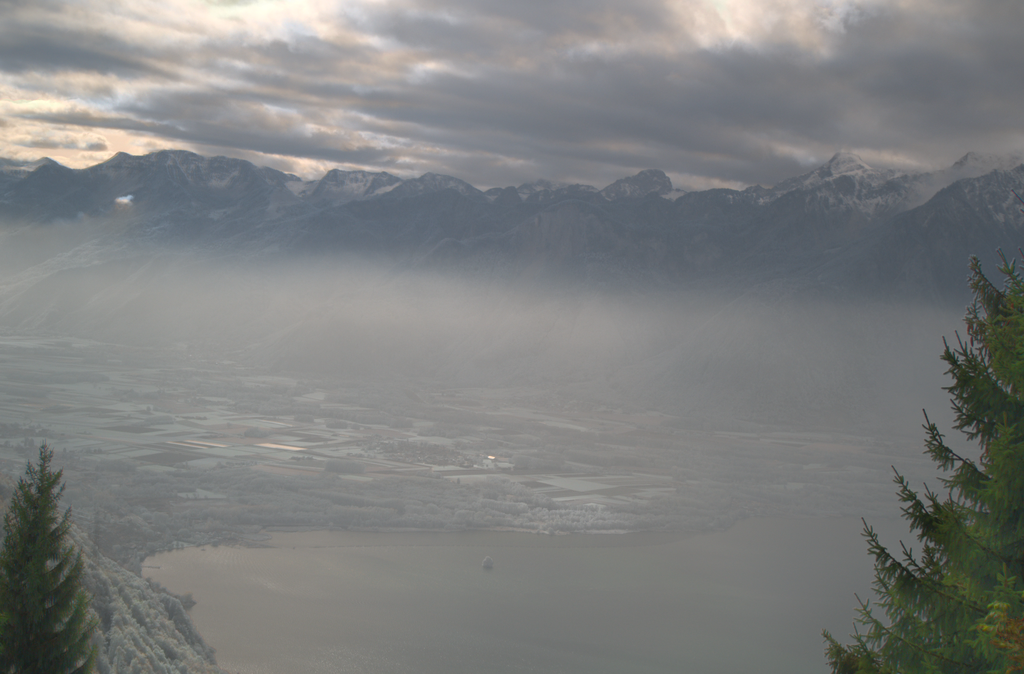}
	\includegraphics[width=80px]{deep_fusionRGB55.png}\\ [0.1cm]
	\includegraphics[width=80px]{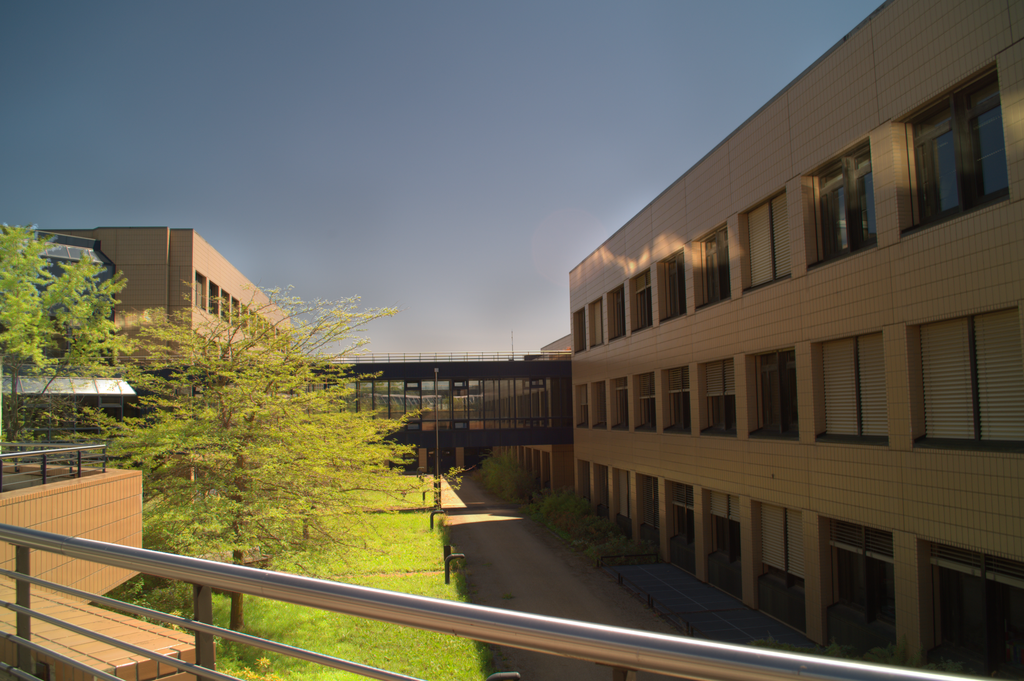}
	\includegraphics[width=80px]{deep_fusionRGB35.png}\\ [0.1cm]
	\includegraphics[width=80px]{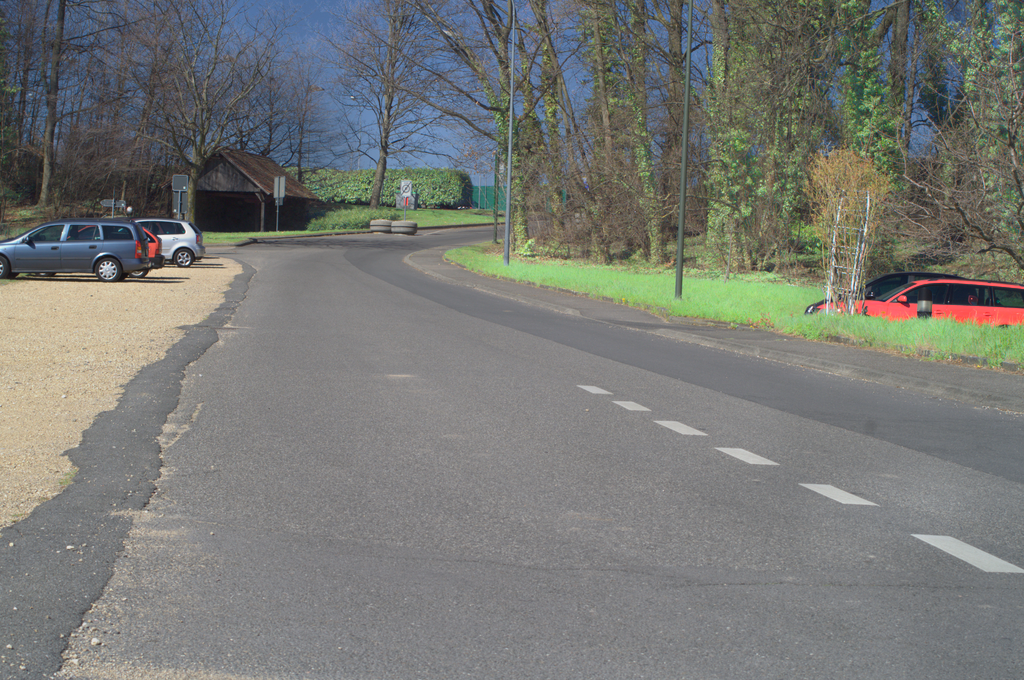}
	\includegraphics[width=80px]{deep_fusionRGB33.png}
 
	\caption{Left: Superpixel image fusion based on classic computer vision \cite{ofir2023multispectral}. Right: the proposed image fusion. From top to bottom: images from the 'Country', 'Mountain', 'Country', 'Urban', and 'Street' categories of the VIS-NIR dataset.}
	\label{fig:fusion_compare}
\end{figure}

\begin{table*}[htb]
	\centering
	\begin{tabular}{| l | c | c| c | c| c | }
        \hline
		Category & SuperPixel \cite{ofir2023multispectral} & PCA \cite{kumar2006pca}& Spectral \cite{ofir2018registration} & SSIM Loss & SSIM+EP Loss\\  
		\hline
		Country & 81.5 & 74.1 & 78.9 & 82.2 & \textbf{82.3}\\
		\hline
		Mountain & 89.9 & 89.6 & 88.1 & 90.1 & \textbf{90.2} \\  
		\hline
        Urban & 93.8 & \textbf{93.9} & 92.7 & 93.4 & 93.5 \\
		\hline
        Street & 87.5 & 87.3 & 85.5 & 87.7 & \textbf{87.8}\\
		\hline
	\end{tabular}
	\caption{Structure of similarity scores of fused images generated by approaches that do not rely on heavy training and/or large datasets.}
	\label{table:metricSSIM}
\end{table*}

\begin{table*}[htb]
	\centering
	\begin{tabular}{| l | c | c| c | c| c | }
        \hline
		Category & SuperPixel \cite{ofir2023multispectral} & PCA \cite{kumar2006pca} & Spectral \cite{ofir2018registration} & SSIM Loss & SSIM+EP Loss \\  
		\hline
		Country & 75.9 & 66.1 & 76.9 & 77.2 & \textbf{77.4} \\
		\hline
		Mountain & 92.7 & 91.0 & 92.8 & 92.9 & \textbf{93.0} \\  
		\hline
        Urban & 95.7 & 95.8 & 95.7 & \textbf{95.9} & \textbf{95.9} \\
		\hline
        Street & 86.9 & 85.0 & 87.1 & 87.6 & \textbf{87.7} \\
		\hline
	\end{tabular}
	\caption{Correlation scores of fused images generated by approaches that do not rely on heavy training and/or large datasets.}
	\label{table:metricCorr}
\end{table*}

The proposed method is evaluated both quantitatively and qualitatively using the multispectral VIS-NIR dataset \cite{BS11}. This dataset is particularly relevant as it offers a variety of scenes captured under different conditions representing the existing fusion challenges in natural images. It contains 954 pairs of NIR and RGB images, divided into four categories, i.e., country, mountain, urban, and street. The dimensions of a typical image from this dataset is 900x768 pixels. 

Quantitative evaluation of the fused images produced by the proposed approach, using either SSIM alone or SSIM and EP as loss functions, is performed using SSIM, Canny \cite{canny1986computational} edge preservation, and statistic correlation. Results are compared to alternative fast methods for image fusion, i.e., SuperPixel \cite{ofir2023multispectral}, PCA Fusion \cite{kumar2006pca}, and Spectral Fusion \cite{ofir2018registration}.

In all reported experiments, the proposed network was trained using 300 epochs (MaxEpoch=300) as it was observed that it is the minimum number of epochs required to produce high-quality fusion images. In terms of hardware, a single standard two-years-old GPU - Nvidia GeForce GTX 3060 - was used leading to fused images being calculated in around several seconds. However, as GPUs have made remarkable progress in terms of processing power since then, usage of a recent GPU model would reduce this computational time to a few seconds.  

Figures  \ref{fig:fusion} and \ref{fig:fusion_results} illustrate visually the quality of the fused  RGB and IR images produced by the proposed method. For all image categories, each channel pair is fused by the same SSL algorithm maintaining relevant information from each spectral channel: edges and details are preserved and relevant color information extracted from the RGB image is present, producing fused images with natural-looking colors.

Further qualitative evidence is provided in Figure \ref{fig:fusion_compare} where fused images generated by the proposed algorithm and those produced by the recent SuperPixel \cite{ofir2023multispectral} method are juxtaposed. Note that none of these methods requires labeling, annotation, or large training datasets. Although the SuperPixel approach is based on classic computer vision meticulously engineered to produce such realistic fused images, the novel approach achieves similar quality having been holistically trained in an end-to-end fashion using a single example.

In terms of quantitative evaluation, Table \ref{table:metricSSIM} and Table \ref{table:metricCorr} compare the proposed algorithm with and without using the EP Loss against other approaches that do not rely on heavy training and/or large dataset by calculating their achieved SSIM fusion and correlation scores, respectively. Whereas the SSIM fusion score is expressed by 
$SSIM(I_1,I_2, F) = 0.5SSIM(I_1, F)+0.5SSIM(I_2,F)$, the correlation metric is defined by
$corr(I_1,I_2, F) = 0.5corr(I_1, F)+0.5coor(I_2,F)$.
Both tables show that the proposed self-supervised fusion approach achieves the best perceptual quality (SSIM) and preserves the best correlation between the input spectral images and their fusion. 
Moreover, these tables highlight the value of integrating edge preservation loss with SSIM in the loss functions for training the CNN. 

This is further confirmed by results shown in Table \ref{table:metric} where edge preservation is calculated for the method having been trained with and without EP loss. For input images $I_1$ and $I_2$, their fusion $F$ and their corresponding Canny \cite{canny1986computational} binary-edges $C_1, C_2,$ and $C_F$, the preservation loss is defined by:
$EP(I_1,I_2) = 0.5\sum_i\frac{\sum_x C_i(x) \cdot C_F(x)}{\sum_x C_i(x)}$.
Table \ref{table:metric} clearly shows that the EP loss enhances significantly the preservation of the edge maps in the proposed self-supervised fusion.

\begin{table}[htb]
	\centering
	\begin{tabular}{| l | c | c |}
        \hline
        Category & SSIM Loss & SSIM+EP Loss \\
        \hline
		Country & 53.5 & \textbf{53.6}\\
		\hline
		Mountain & 55.9 & \textbf{56.6} \\
		\hline
        Urban & 75.2 & \textbf{75.5} \\
		\hline
        Street & 58.0 & \textbf{58.5} \\
		\hline
	\end{tabular}
	\caption{Percentage of initial Canny edges remaining in the fused images for the proposed method having been trained with and without EP loss.}
	\label{table:metric}
\end{table}

\begin{table}[htb]
	\centering
	\begin{tabular}{| l | c | c| c | }
        \hline
		Category & Compact & C.+UNet & C.+UNet+STN \\  
	   \hline
		Parameters & $\approx 1400 $& $\approx 14M$ & $\approx 27M$  \\	
        \hline
        Runtime & $\approx 10s$ & $\approx 40s$ & $\approx 40s$  \\	
        \hline
		SIMM score & 86.3 & \textbf{86.5} & 83.0  \\
		\hline
	\end{tabular}
	\caption{Ablation study of the proposed architecture assessing the contribution to the compact CNN architecture (or ’Compact’, or 'C.') of the fusion map produced by the UNet-Resnet18 component (or ’Unet’) and the cost of conducting image alignment using STN.}
	\label{table:ablation}
\end{table}

Finally, outcomes of an ablation study is reported in Table \ref{table:ablation} to assess the contribution to the compact CNN architecture (see Figure \ref{fig:compact}) of the fusion map produced by the UNet-Resnet18 component and the cost of conducting image alignment using STN. In terms of the SSIM score obtained by processing 10\% of the whole VIS-NIR dataset, although the compact CNN can already fuse the input images with high quality, the inclusion of the UNet-Resnet18 parallel results in a modest, but clear performance improvement. However, this is associated to processing time being quadrupled due to the major increase of learnable parameters. This study also reveals that usage of the off-the-shelf STN to deal with cases of misregistration between the spectral channels affects quite dramatically fusion quality. However, in such situations, this is an unavoidable compromise that needs to be made.

Overall, experiments conducted in this section demonstrate that the proposed self-supervised fusion method trained on a single example achieves high-quality image fusion and state-of-the-art performance when evaluated against comparable approaches.

\section{Conclusions} \label{sec:conclusions}

This paper introduces a novel and practical approach for infrared and visible image fusion based on fast training of a self-supervised short CNN relying on a single example pair of images. Its technical contributions, highlighted by an ablation study, include the addition of a parallel stream computing accurate fusion map to enhance quality and a training loop algorithm relying on both SSIM and EP loss functions. 

In terms of fast multispectral fusion, experiments have shown that the proposed method delivers similar or better results both quantitatively and qualitatively than alternative approaches that do not rely on heavy training and/or large datasets.


	{\small
		\bibliographystyle{ieee}
		\bibliography{egbib}
	}
	
\end{document}